\documentclass[letterpaper, 10 pt, conference]{ieeeconf}  % Comment this line out if you need a4paper

\IEEEoverridecommandlockouts                              % This command is only needed if 
\usepackage{graphics} % for pdf, bitmapped graphics files
\usepackage{epsfig} % for postscript graphics files
\usepackage{mathptmx} % assumes new font selection scheme installed
\usepackage{times} % assumes new font selection scheme installed
\usepackage{amsmath} % assumes amsmath package installed
\usepackage{amssymb}  % assumes amsmath package installed
\usepackage{multirow}
\usepackage{url}
\usepackage[noadjust]{cite}
\usepackage{color}

\title{\LARGE \bf
Spatial-Temporal Mitosis Detection in Phase-Contrast Microscopy via Likelihood Map Estimation by 3DCNN
}

\author{Kazuya Nishimura$^{1}$ and Ryoma Bise$^{1}$% <-this % stops a space
\thanks{$^{1}$Department of Advanced Information Technology, Kyushu University, Fukuoka, Japan
}
\thanks{    
{\tt\small kazuya.nishimura@human.ait.kyushu-u.ac.jp}}%
}

\begin{document}

\maketitle
\thispagestyle{empty}
\pagestyle{empty}

%%%%%%%%%%%%%%%%%%%%%%%%%%%%%%%%%%%%%%%%%%%%%%%%%%%%%%%%%%%%%%%%%%%%%%%%%%%%%%%%
\begin{abstract}
Automated mitotic detection in time-lapse phase-contrast microscopy provides us much information for cell behavior analysis, and thus several mitosis detection methods have been proposed. However, these methods still have two problems; 1) they cannot detect multiple mitosis events when there are closely placed. 2) they do not consider the annotation gaps, which may occur since the appearances of mitosis cells are very similar before and after the annotated frame. In this paper, we propose a novel mitosis detection method that can detect multiple mitosis events in a candidate sequence and mitigate the human annotation gap via estimating spatial-temporal likelihood map by 3DCNN. In this training, the loss gradually decreases with the gap size between ground-truth and estimation. This mitigates the annotation gaps. Our method outperformed the compared methods in terms of F1-score using challenging dataset that contains the data under four different conditions.
Code is publicly available in \url{https://github.com/naivete5656/MDMLM}.
\end{abstract}

%%%%%%%%%%%%%%%%%%%%%%%%%%%%%%%%%%%%%%%%%%%%%%%%%%%%%%%%%%%%%%%%%%%%%%%%%%%%%%%%

\section{INTRODUCTION}
% 生きた細胞の解析は有用で非侵襲な位相差像が使用される．
Living cell analysis has an important role in biomedical research such as investigating the effect of a drug and the analysis of cell fate. 
To visualize living cell, phase-contrast microscopy, which is a non-invasive technique, has been widely used for long term monitoring cell populations in vitro. 
Unlike other invasive imaging methods such as fluorescent assays, it allows cell observation without cell staining.

% 細胞分裂の解析は増殖の解析ができる．手動解析が大変だから，自動解析が必要．
The mitosis detection in phase-contrast microscopy provides much information such as the proliferative behavior of the cell population under specific cultured conditions. 
It is also expected to improve the automated cell tracking methods~\cite{huh2011mitosis, yang2005cell}.
% As shown in Fig.~\ref{fig:introduction}, the cell appearance changes depends on cultured conditions. The proliferative behavior is also change.
The manual analysis is time-consuming, tedious, and prone to human error for a large amount of data.
Therefore, the automated mitosis detection method in the phase-contrast image is necessary and it provides us quantitative information on cell proliferation.

\begin{figure}[t]
    \centering
    \includegraphics[width=\linewidth]{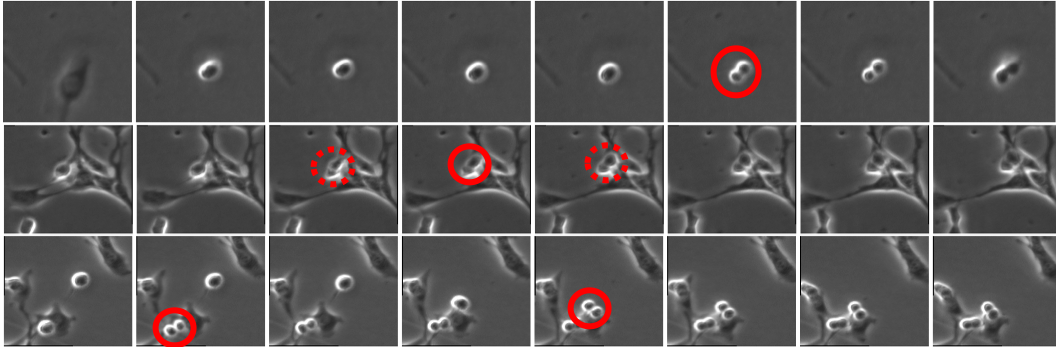}
    \caption{Example sequences of mitosis process. Red circles indicate human annotations where and when the mitosis event occurs.
\textbf{Top}: an easy case that the single cell becomes shrink and bright and then divided into two cells. The timing is clear.
\textbf{Middle}: a difficult case that the appearance of the annotated cell (red circle) is similar to those at before and after the frame (red dotted circles). 
\textbf{Bottom}: a case that multiple mitosis events occur in a near position.}
    \label{fig:introduction}
\end{figure}

% 細胞分裂の説明. こういう振る舞いを持って，こういう特性がある．
The top row in Fig.~\ref{fig:introduction} shows an example process of mitosis. In this process of mitosis, the circularity and pixel value around the mitosis cell becomes high. Then the cell is splitting into two daughter cells, gradually expanding and return to normal cells. 
Our aim of this study automatically detects the coordinate $x, y$ and frame $t$ of the moment that the two daughter cells first appeared and the boundary of two cells can be observed.

To detect mitosis events, several methods have been proposed.
Almost all methods first extract the candidate sequences, and these methods identify the frame of the mitosis event from the candidate sequence via graphical model~\cite{huh2011automated, liu2010automated}, CNN~\cite{nie20163d, mao2016hierarchical}, or LSTM~\cite{mao2019cell, su2017spatiotemporal} to recognize the temporal feature change.
% these methods estimate the mitosis event frame with a graphical model~\cite{huh2011automated, liu2010automated,huh2011detection, gallardo2004mitotic}, CNN~\cite{nie20163d, mao2016hierarchical} or LSTM~\cite{su2017spatiotemporal} that is trained to recognize the temporal feature change.
% That appears the spatial localization (i.e. mitosis event position) relies on the candidate extraction that does not use temporal information, and the temporal information is only used for temporal localization (i.e. mitosis event timing).
% \begin{figure}[t]
%     \centering
%     \includegraphics[width=\linewidth]{figures/Intro_figure.png}
%     \caption{(a), (b), (c), (d) is the example of a phase-contrast image that is cultured on various conditions. 
%     Each condition is cultured on each growth factor: (a) Control, (b) FGF2, (c) BMP2, (d) FGF2+BMP2 respectively.
%     The red circles indicate mitosis events and the blue circles indicate non-mitosis cells that look like mitosis events. 
%     (e) The example process of mitosis. Positive and negative example means the sequence that includes and does not include cell mitosis.}
%     \label{fig:introduction}
% \end{figure}

However, these methods still have two problems. 
First, these methods can not detect multiple mitosis events in a candidate sequence since they assume that a candidate sequence contains only one mitosis event. This situation may occur when the mitosis positions are close as shown in the third row in Fig.~\ref{fig:introduction}.
In addition, the spatial localization (i.e. mitosis event position) relies on the candidate extraction and it does not use temporal information. The temporal information is only used for temporal localization (i.e. mitosis event timing).

Second, as shown in the red circle and red dotted circle in the middle row in Fig.~\ref{fig:introduction}, the appearance of the mitosis cell looks the same before and after the frame.
Therefore, the annotations by different annotators may have a temporal gap, and it is difficult for even an expert to annotate with consistent criteria, {\it i.e.}, it may contain gaps in time. 
The existing methods estimate a single frame that seems occurs mitosis event from the candidate sequences. 
If the estimated frame is very close from the ground-truth, the estimate is almost correct and it is different from a miss estimation.
Nevertheless, the previous approaches do not consider this situation and it is treated as a miss estimation.

% Second, when the mitosis events occur in close positions such as shown in the third row in Fig. \ref{fig:introduction}, these methods can not detect two mitosis events.
% The candidate extraction of almost methods is based on pixel value intensity thus the sequences sometimes include two mitosis events if the mitosis position is near. 
% Despite, these methods assume that the candidate sequence includes one mitosis event, thereby these methods can not detect all mitosis events in the candidate sequence including some mitosis events.
% In addition, the spatial localization ({\it i.e.} mitosis event position) relies on the candidate extraction that does not use temporal information, and the temporal information is only used for temporal localization ({\it i.e.} mitosis event timing).
\begin{figure*}[t]
    \centering
    \includegraphics[width=\linewidth]{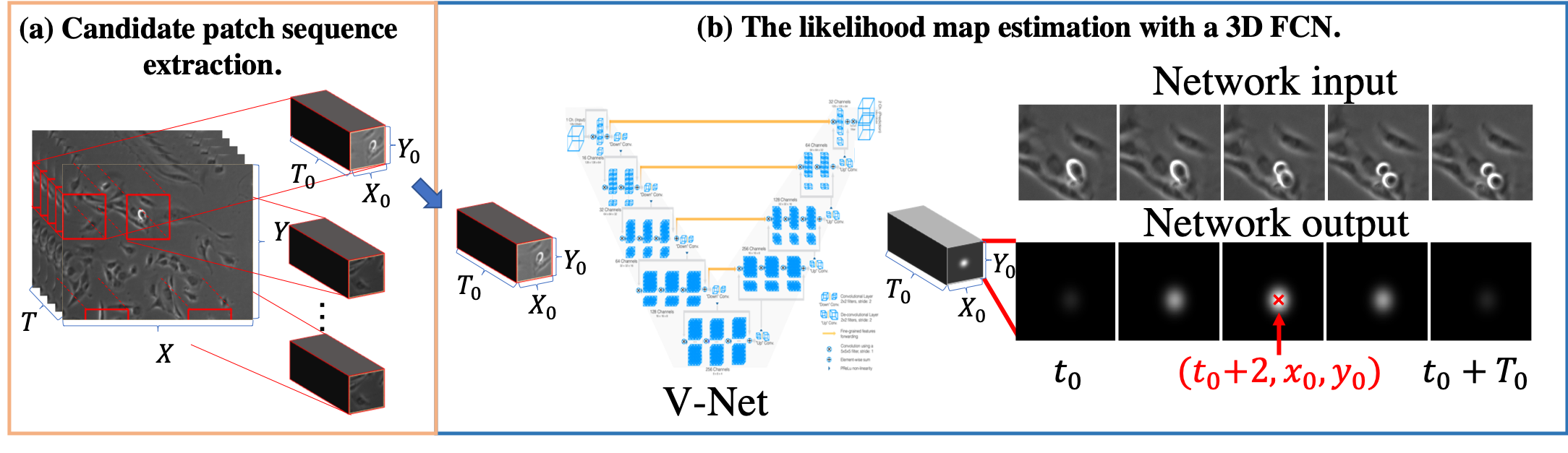}
    % \caption{手法の概要}
    \caption{The overview of the proposed method. (a) First, candidate sequences are extracted based on the fact that the pixel value becomes high in the process of mitosis. b) Then, mitosis events are detected from each candidate via estimating the likelihood map of cell mitosis events by using V-Net \cite{vnet2016v}. The peak point in the estimated 3D map indicates the mitosis position and timing.}
    \label{fig:overviewmethod}
\end{figure*}
\textbf{Contribution}:
The main contribution of this work is to propose a novel mitosis detection method that can detect multiple mitosis events in a candidate sequence and mitigate the human annotation gap.
Unlike the previous models that only localize temporal timing, our method can localize the spatio-temporal locations of mitosis events from a candidate sequence.
Our model estimates the 3D (2D+t) likelihood map of mitosis events as shown in the network output in Fig.2. This likelihood map can represent multiple mitosis events in a 3D volume (a candidate sequence), and thus it can detect multiple mitosis events. 
% Our model predicts the cell mitosis likelihood map (network output in Fig.~\ref{fig:overviewmethod}.) for each pixel of 3D (2D + temporal) input thus it is possible to deal with multiple mitosis events inside the candidate sequence.
In addition,  we effectively use temporal information for spatial localization by using 3D (2D+t) convolution.
Our likelihood map estimation is effective for reducing the affect from the human annotation gap, in which the annotated coordinates become peak with a Gaussian distribution.
In the training of our model for estimating the likelihood map, a small gap of spatial-temporal localization gives a small loss, and thus this mitigates the gaps.
In experiments, we evaluated our method using a challenging dataset~\cite{eom2018phase} that was used in the CVPR Workshop contest~\cite{cvprmc2019} and we outperformed the other compared methods. 
Our proposed method won second place in the CVPR Workshop contest~\cite{cvprmc2019}.

\textbf{Related work:}
The tracking-based mitosis detection methods first track all cells and then the location of a newly appeared cell in the tracking result is extracted as a candidate for mitosis events. Then, the candidate is classified using visual appearance~\cite{thirusittampalam2013novel,al2006automated,bise2009reliably, yang2005cell}.
In this approach, mitosis detection relies on cell tracking results. However, the tracking performance significantly decreases when the cell density is high. In addition, the temporal information is not used for classifying the detected candidates if it is true or not. 
When multiple cells touch each other, the appearance of the touched cells is similar to a mitosis cell. 
It is difficult to recognize mitosis without temporal information.
Therefore, these methods do not achieve enough performance.

As discussed above, the appearance change of a mitosis cell is important information for mitosis detection. To effectively use the temporal context, several graphical model-based mitosis detection methods that model the change of the visual features in time have been proposed.
These methods first extract candidate sequences from the original sequence using image processing. Next, the visual feature of each patch image is extracted using hand-crafted feature extraction such as SIFT \cite{lowe2004distinctive}.
Then the graphical model such as HMM~\cite{gallardo2004mitotic} or EDCRF~\cite{huh2011automated} or MGRF~\cite{liu2017multi} MM-HCRF+
MM-SMM~\cite{liu2012semi} is used to recognize the mitosis frame from the candidate sequence.
These methods achieved better performance than the tracking-based method by benefit from time-series information.

Recently, deep learning is used in the same framework with the graphical model-based method.
Nie {\it et al.}~\cite{nie20163d}  used 3DCNN for extracting candidate regions and classified by SVM whether the sequence contains mitosis. However, this method does not have the ability to localize when the mitosis occurs in the sequence.
Zhou {\it et al.}~\cite{zhou2017cell} proposed a method that estimates the score map by 2D CNN, in which the map indicates the candidate of mitosis event on each image by 2D CNN and it is classified by 3D CNN whether the sequence includes mitosis. However, the input of 3D CNN  does not contain cell appearance information, thereby the method can not use the feature extraction ability of 3D CNN.
Mao {\it et al.}~\cite{mao2016hierarchical, mao2019cell} proposed a method that models the appearance changes of a mitosis cell using a bidirectional LSTM for identifying the timing of a mitosis event. 
Su {\it et al.}~\cite{su2017spatiotemporal} proposed a model that uses CNN and LSTM for mitosis detection. The CNN extracts the spatial feature and the LSTM extracts the temporal feature.
These methods improved the performance compared with those using the hand-crafted feature extraction.
However, they assume that a mitosis event occurs one time in the candidate sequence, and thus it is difficult to apply it to the data under the dense condition.
In addition, these methods do not deal with the human annotation gap. Sometimes the human may annotate before or after frame that the network estimates.
This affects the network training.

To address these problems with multiple mitosis detection and annotation gaps, our method estimates the likelihood map that can represent multiple mitosis events and reduce the bad effects of a human annotation gap.

\section{Method}
Fig.~\ref{fig:overviewmethod} shows the overview of our proposed method.
Our method consists of following two steps.
\begin{enumerate}
\renewcommand{\theenumi}{\alph{enumi}}
\item \textbf{Candidate patch sequence extraction:}

In the entire image sequence, mitosis events appear very sparsely, and thus it occurs the imbalance of true positives and negatives in detection.
Learning a CNN with an imbalance dataset takes a lot of time, and it is difficult to converge the training.
To mitigate a data imbalance, we first extract candidate sequences that contain the mitosis events by using the intensity-based method based on the fact that the intensity around the cell becomes high when a mitosis event occurs.

\item \textbf{The likelihood map estimation with a 3D FCN:}

Multiple mitosis events may occur in a candidate sequence. Therefore, we handle this problem as a detection problem rather than a classification problem.
We detect a mitosis event by estimating the cell mitosis likelihood map for the candidate sequence. 
To estimate the likelihood map, we use 3D FCN that can estimate each pixel for an input image.
In the likelihood map, a mitosis position is represented as an intensity peak with a Gaussian distribution, in which multiple mitosis are represented as multiple peaks.
% 位置ピクセル毎の分裂確率
% the method can localize the
% Spatio-temporal locations of mitosis events from a candidate
% sequence. Our model predicts the cell mitosis likelihood map
% for each pixel of 3D (2D + temporal) input thus it is possible
% to deal with multiple mitosis events inside the candidate
% sequence. The cell mitosis likelihood map is generated as fol-
% Fig. 2. The overview of our proposed method. First, the candidate sequence is ext
% Furthermore, the annotation may be not the same as accurate cell mitosis timing since the human annotations are not generally strict.
% To represent this problem, we proposed an output representation that predicts the cell mitosis likelihood map.
% Since cells often touch with each other and the touching cells look similar to the appearance of cells right after a mitosis event, it is difficult to identify mitosis events only using a single time-frame. 
% It is known that a cell under mitosis becomes increasing circularity and brightness, and then gradually spreads the cell region.
% To identify this mitosis event, the time-series appearance change information is therefore important.
% We handle this time-series information using a 3D convolutional neural network, where the time frame is used as z-slice.

\end{enumerate}

\subsection{Candidate patch sequence extraction}
We follow the previous work proposed by Huh \cite{huh2011automated} for candidate patch sequence extraction.
Since it is known that the process of mitosis event typically exhibits high brightness, the patch images ({\it i.e.} the candidate images that may include some mitosis event.) can be extracted based on the brightness~\cite{huh2011automated}.
Each original image is convolved with a small-sized $d \times d$ rectangular average filter.
Then the image is binarized by thresholding and the bounding box that has a high probability of including mitosis is obtained from the connected component of the binarized image.
Next, the bounding boxes are associated with successive frames based on the distance of the boxes and the candidate sequence is generated (Fig.~\ref{fig:overviewmethod} (a)).
We treat the candidate sequence as a 3D image by considering the time-series as z-slice. It is used the network input.
This candidate extraction mitigates the imbalance problem of detection since many regions of false negatives ({\it i.e.}, non-mitosis regions) are excluded.
% Fig. 3 shows examples of candidates of mitosis processes, in which a red box shows the moment and position of the mitosis event.
% Fig.~\ref{fig:example_sequence} shows the example of an extract candidate sequence.

% \begin{figure}[t]
%     \centering
%     \includegraphics[width=\linewidth]{./figures/candidate_example.png}
%     \caption{The example of candidate sequence.}
%     \label{fig:example_sequence}
% \end{figure}

\subsection{The likelihood map estimation with a 3D FCN}
%Since the appearance of this moment is similar to that of the other timing, the time series information is necessary to detect the timing of cell division.
%Therefore, we use the 3D convolutional neural network (V-Net~\cite{vnet2016v}) that can extract the temporal feature by using the z-axis as the temporal axis.
To localize the spatial-temporal positions of mitosis events with effectively using the spatial-temporal information, we use a 3D fully convolutional neural network (V-Net~\cite{vnet2016v}), which has been used for 3D segmentation problem. This network can extract the temporal feature by using the z-axis as a temporal axis.

Furthermore, the annotation may be off the accurate cell mitosis timing since the human annotations are not generally strict. 
To deal with the gap between the human annotation and the accurate cell mitosis position, we use the cell mitosis position likelihood map as training data, where an annotated position becomes a peak and the value gradually decreases with a Gaussian distribution.

The network input is the candidate sequence and the output is the cell mitosis likelihood map which is the same size as the input and the local maxima of output becomes mitosis event position (Fig.~\ref{fig:overviewmethod} (b)).
% To train V-Net for detecting cell mitosis, we generate 3D image for the network input using the candidate sequence. 
% As shown in Fig.\ref{fig:overviewmethod}, 
% The inputs and outputs of V-Net are candidate sequences and cell mitosis likelihood map that . 
% The inputs and outputs of V-Net are candidate sequences and the cell mitosis likelihood maps.
% The V-Net output the cell likelihood map
% The each pixel of the cell mitosis likelihood map indicates mitosis likelihood. The range of output is 0 to 1. 
% V-Net can predict each pixel for the input 3D image.
% V-Net is trained and used to output a cell mitosis likelihood map.

We generate the cell mitosis likelihood map $L$ to be used as ground-truth at each candidate sequence $i\in{N}$. $N$ is the total number of the candidate sequence.
To deal with the case that the candidate sequence includes multiple mitosis events, the likelihood map is generated for each annotation in the candidate sequence and these are aggregated by the max operation. 
Let $\mathbf{p} \in \mathbb{R}^3$ denote a location of the likelihood map, and $p_x, p_y, p_t$ are the coordinates of each axis.
The likelihood map for the $k$-th annotation $(x_k, y_k, t_k)$ is defined as, 
\begin{equation}
    L^k_i ( \mathbf { p }) = \exp \left( - \left( { \frac{(p_x - x_ k) ^ 2}{\sigma _x^2} + \frac{(p_y - y_ k) ^ 2}{\sigma _y^2} + \frac{(p_t - t_ k) ^ 2 }{\sigma _t^2}}\right) \right),
\end{equation}
where $\sigma_x,~\sigma_y,~\sigma_z$ are the hyper parameters which control the spread of a peak..
The likelihood maps $L^k (\mathbf{ p })$ ($k=1,...,M$) for each annotation are aggregated by a max operation.
\begin{equation}\label{eq:eachlikely}
    {L}_i ( \mathbf { p } ) =  \max_k( L^k_i (\mathbf{ p })).
\end{equation}
We use the mean of the squared error loss function (MSE):
\begin{equation}
    Loss = \frac{1}{N} \sum^{N}_{i=1}||L_i - O_i||^2_2,
\end{equation}
where $O_i$ is the network output.
We execute a backpropagation algorithm of the computed loss to update the network parameters.
In this training, the loss gradually decreases with the gap size between the ground-truth and estimation. It indicates that a small gap gives a small loss in contrast to the previous methods that give the same loss with that of the large gap even for the small gap.
This mitigates the annotation gaps.

\section{Experiments}

% \begin{figure}[t]
%     \centering
%     \includegraphics[width=\linewidth]{figures/dataset.pdf}
%     \caption{The image that cultured each condition.}
%     \label{fig:dataset}
% \end{figure}
\subsection{Dataset}
\begin{figure}[t]
    \centering
    \includegraphics[width=\linewidth]{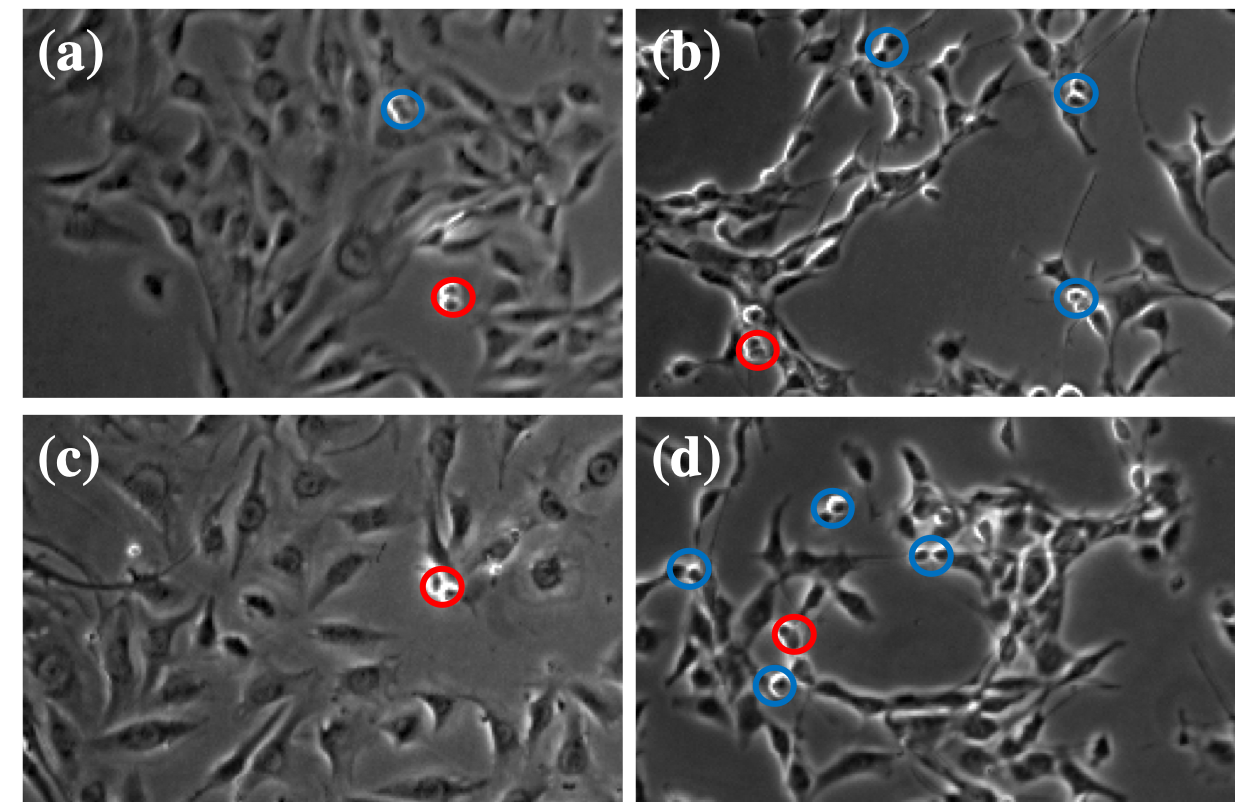}
    \caption{Example images under various conditions where cells were cultured under growth factors: (a) Control, (b) FGF2, (c) BMP2, (d) FGF2+BMP2 respectively. Red circle indicates a mitosis cell and blue indicates a non-mitosis cell that looks similar to a mitosis event. (b), (d)  contain many blue circles hence these are difficult conditions.}
    \label{fig:Dataset}
\end{figure}

% The sequences were captured stem cells under 4 different conditions (Control, FGF2, BMP2, FGF2+BMP2) by a phase-contrast microscope.
% Each sequence has the annotation of the mitosis events (t, x, y). t is the frame number of mitosis position and x, y is the spatial position \footnote{https://media.m2i.ac.cn/mitosisdetection/evaluation/}
We evaluated our method using a challenging open dataset~\cite{eom2018phase} that was used in the CVPR2019 mitosis detection contest~\cite{cvprmc2019}.
The dataset contains 12 sequences of time-lapse images of the myoblast stem cell that were captured by phase-contrast microscopy\footnote{The original dataset contains 12 sequences for training and 4 for the test set. The training data is only opened and the test data is closed.~\cite{cvprmc2019}}.
The images were captured every 5 minutes and each sequence consists of 1013 images with a resolution of 1392$\times$1040 pixels.
The sequences are divided into four groups depending on the growth factor; (a) Control, (b) FGF2, (c) BMP2, (d) FGF2+BMP2~(Fig~\ref{fig:Dataset}. (a), (b), (c), (d) show the examples of each condition.).
There are three sequences for each condition.
The human annotation of mitosis events (t, x, y) was given for each sequences.
$t$ indicates the time frame, and $x,~y$ indicate the spatial coordinate of a mitosis event.
In the condition (b) and (d), the appearances of the cells are similar to mitosis cells, {\it i.e.,} their appearance becomes high intensity and thus it is more difficult to detect mitosis events in these conditions compared with (a), (c).

The V-Net was trained on each cultured condition.
We used two sequences as the training data and the other one as the test data.
The network input (z is a temporal axis) is generated by cropping from the center of the first frame of the candidate sequence. 
The input image size of V-Net is $128\times128\times16$ in all experiments.
If the candidate sequence was shorter than 16 frames, 0 paddings were applied. If the candidate was longer than 16, the sequence was separated into several candidates by sliding window, where we set the value of the slide as 8.

% \subsection{Experiment condition}
% The input image is always $128\times 128 \times 16$.
% The candidate is shorter than 16 then 0 paddings apply.
% The candidate is longer than 16 then the sequence is cut by sliding window with stride 8.
% バッチサイズは10，損失関数は平均二乗誤差，最適化手法はadam（learning rateは1e-3）,エポック数は50としエポック中で最もvalidation lossが低い重みを使用する．

\subsection{Detection Accuracy.}

% \begin{figure}[t]
%     \centering
%     \includegraphics[width=\linewidth]{figures/results.pdf}
%     \caption{The example results of dense cell density condition. The red points indicate true positive results, and blue points indicate false-positive results.}
%     \label{fig:result}
% \end{figure}

In the experiments, we compared our method with two other methods and one modified our method for ablation study; 1) 3DCNN \cite{nie20163d} that classifies whether the candidate sequence contains a mitosis event, 2) EDCRF~\cite{huh2011automated} that uses a CRF to identify the mitosis timing from a candidate sequence. In addition, to show the effect of our regression setup, in which the network produces the likelihood map rather than segmentation, we also evaluted 3) ours w/o cell mitosis likelihood map (CMLM) that uses the same V-Net with the proposed method but it was trained to produce the segmentation results, where the mitosis region is 1, otherwise 0.
Since 3DCNN~\cite{nie20163d} does not have temporal localization ability, we set the center coordinate of an input sequence as the timing of the mitosis event.

We used the precision, recall, and F1-score that were used in the contest~\cite{cvprmc2019}. 
If a detected position sufficiently closes to that of a ground-truth (within 6 frames in the temporal direction and 15 pixels in the spatial direction\footnote{We followed the contest \cite{cvprmc2019} to set these values.}), it is counted as a true positive and if it is far from a ground-truth, it is counted as false positive.
% The detection result sufficiently close to the ground-truth (within 6 frames in the temporal direction and 15 pixels in the spatial direction) define as a truly positive and far result as false positive. 
If there is no detection points in the local window form a ground-truth, it is counted as a false negative.
\begin{figure}[t]
    \centering
    \includegraphics[width=\linewidth]{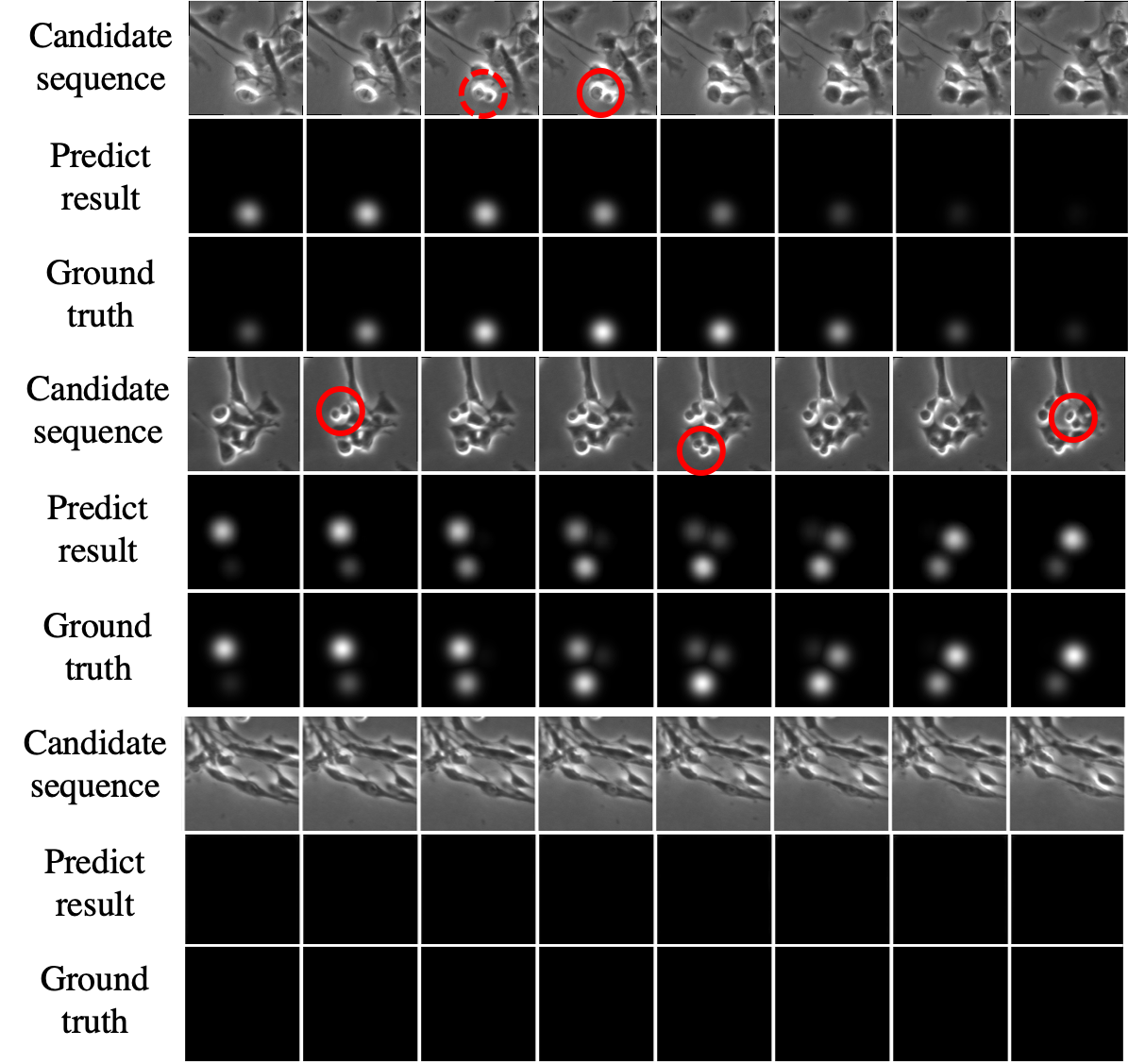}
    \caption{Examples of mitosis detection results in difficult cases.
\textbf{Top}: the sequence that the appearance on the annotated mitosis timing (red circle) is similar to that on the previous frame (dotted red circle). 
\textbf{Middle}: the sequence that contains multiple mitosis events. 
\textbf{Bottom}:  the negative candidate sequence that contains no mitosis events.}
    \label{fig:example_result}
\end{figure}

\begin{table*}[t]
    \centering
    \caption{Comparison of mitosis detection.}
    \label{fig:mitosis}
    \begin{tabular}{c|c|c|c|c|c|c|c|c|c|c|c|c}
    \hline
     & \multicolumn{3}{c|}{Control} & \multicolumn{3}{c|}{FGF2} & \multicolumn{3}{c|}{BMP2} & \multicolumn{3}{c}{FGF2+BMP2}  \\  \cline{2-13}
                  & Precision   & Recall   & F1  & Precision   & Recall   & F1 & Precision   & Recall   & F1 & Precision   & Recall   & F1   \\ \hline \hline
    3DCNN~\cite{nie20163d} &0.556& 0.403 & 0.467&0.345& 0.290& 0.315&0.500&0.418 & 0.456& 0.345& 0.285& 0.312\\ 
    % 3DCNN + RT~\cite{nie20163d} &0.767& 0.743& 0.755& 0.494& 0.730& 0.0.589& 0.823& 0.925& 0.871& 0.580& 0.731& 0.647\\ \hline
    Huh~\cite{huh2011automated} &0.699& 0.765& 0.731& 0.347& 0.454& 0.394& 0.840& 0.845& 0.843& 0.539& 0.604& 0.569\\ \hline
    Ours w/o CMLM&0.813& 0.865& 0.838& 0.345& \textbf{0.875}& 0.495& 0.750& \textbf{0.972}& 0.847& 0.395& \textbf{0.895}& 0.548\\ 
    Ours  &\textbf{0.857}& \textbf{0.898}& \textbf{0.877}& \textbf{0.646}& 0.841& \textbf{0.731}& \textbf{0.864}& 0.947& \textbf{0.904}& \textbf{0.831}& 0.746& \textbf{0.786}\\ \hline
    \end{tabular}
\end{table*}
The performances of each method are shown in Tab.~\ref{fig:mitosis}.
In all conditions, our method outperformed the other methods in terms of F1-score.
In addition, compared with Ours w/o CMLM, our method improved in terms of F1-score. 
This indicates the effectiveness of the proposed likelihood map estimation.

Fig.~\ref{fig:example_result}  shows difficult cases due to the similar appearance with the normal cells.
The red circle is the annotated position and timing in this examples, and, the dotted circle position also looks like the moment of the mitosis event.
As mentioned above, the previous methods treat even one frame gap as a miss estimation in training although the timing may change depending on the annotator.
This affects training.
On the other hand, in our method, the loss for one frame gap is small since our method estimates the likelihood map as shown in the ground-truth in Fig.~\ref{fig:example_result}. Therefore, the adverse effect for training is small and it can achieve robust detection.

In particular, the performance of the compared methods significantly decrease in the condition of FGF2 and FGF2+BMP2, {\it i.e.}, the difficult cases. In these conditions, the cell density is high where cells shrink and have high intensities, and thus multiple mitosis events often occur in a candidate sequence as shown in Fig.~\ref{fig:example_result}, middle. It is difficult for a non-expert to identify all the three mitosis events in the example. In spite of this difficult condition, our method successfully estimated the likelihood map and detected these three mitosis events (Fig.~\ref{fig:example_result}). As a result, our method improved the F1-scores over $20\%$ from those of the other methods in these conditions.

\begin{figure}[t]
  \centering
    \begin{tabular}{c}
      % 1
      \begin{minipage}{0.48\hsize}
        \centering
        \includegraphics[width=\linewidth]{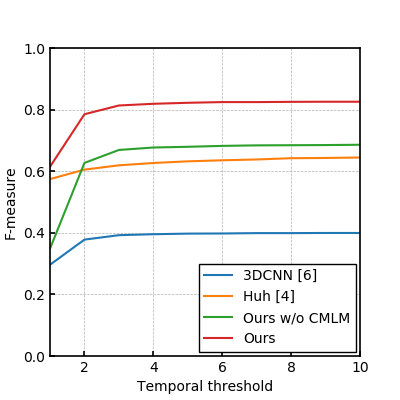}
        \caption{Comparison of mitosis event localization on F1-score with changing the temporal thresholds.}
        \label{fig:thresh_t}
      \end{minipage}
      % 2
      \begin{minipage}{0.48\hsize}
        \centering
        \includegraphics[width=\linewidth]{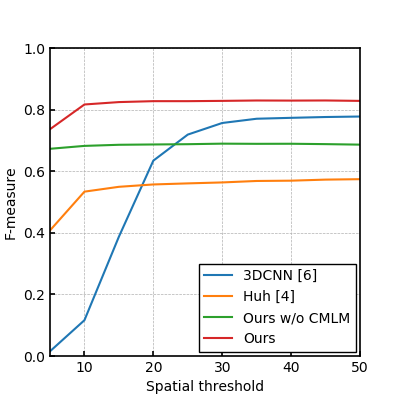}
        \caption{Comparison of mitosis event localization on F1-score with changing the spatial thresholds.}
        \label{fig:thresh_xy}
      \end{minipage}
    \end{tabular}
\end{figure}

\subsection{Evaluation of spatial and temporal localization ability.}
In the above experiments, we used the fixed spatial-temporal thresholds to define true positives of detection.
In order to show the robustness of our method for the spatial-temporal localization, we evaluated the F1-scores with changing the spatial and temporal thresholds, in which Fig.~\ref{fig:thresh_t} shows the results for temporal and Fig.~\ref{fig:thresh_xy} for spatial. Here, the high performance with a small threshold indicates the method can accurately localize the spatial position and the timing.
In Figs.~\ref{fig:thresh_xy} and ~\ref{fig:thresh_t}, the performances of our method were the best at all the thresholds in the comparison. It indicates that our method is effective for both spatial and temporal localization. It is considered for the reason that our method utilizes the spatial-temporal context via 3D convolution for both spatial-temporal localization, in contrast to the other methods that do not use the temporal information for candidate sequence detection (i.e., spatial localization).

\section{Conclusion}
In this paper, we proposed an effective method for spatial-temporal mitosis detection by estimating the cell mitosis likelihood map.
The method can detect multiple mitosis events and mitigate the annotation gap from ground-truth. 
In experiments, we confirmed that our proposed method performs better than other methods.
In addition, we demonstrated the effectiveness of our method for spatial-temporal localization.
The proposed method won second place in the contest of CVPR workshop cell mitosis detection~\cite{cvprmc2019}.

%%%%%%%%%%%%%%%%%%%%%%%%%%%%%%%%%%%%%%%%%%%%%%%%%%%%%%%%%%%%%%%%%%%%%%%%%%%%%%%%

%%%%%%%%%%%%%%%%%%%%%%%%%%%%%%%%%%%%%%%%%%%%%%%%%%%%%%%%%%%%%%%%%%%%%%%%%%%%%%%%



\section*{ACKNOWLEDGMENT}
This work was supported by JSPS KAKENHI Grant Number JP18H05104 and JP19K22895.
% The preferred spelling of the word ÒacknowledgmentÓ in America is without an ÒeÓ after the ÒgÓ. Avoid the stilted expression, ÒOne of us (R. B. G.) thanks . . .Ó  Instead, try ÒR. B. G. thanksÓ. Put sponsor acknowledgments in the unnumbered footnote on the first page.

%%%%%%%%%%%%%%%%%%%%%%%%%%%%%%%%%%%%%%%%%%%%%%%%%%%%%%%%%%%%%%%%%%%%%%%%%%%%%%%%

\newpage

\bibliographystyle{IEEEtran}
\bibliography{IEEEabrv}

\end{document}